\documentclass[letterpaper, 10 pt, conference]{ieeeconf} 
\IEEEoverridecommandlockouts     
\usepackage{cite}
\usepackage{hyperref}
\usepackage[pdftex]{graphicx}
\usepackage{epstopdf}
\usepackage{amsmath}
\usepackage{amssymb}
\usepackage{array}
\usepackage{wasysym}
\usepackage{bm}
\usepackage{units}
\usepackage[per-mode=symbol]{siunitx}
\usepackage{url}
\usepackage{color}
\usepackage{epstopdf}
\usepackage[ruled]{algorithm2e}
\usepackage{multirow}
\usepackage{booktabs}
\usepackage{lineno}
\usepackage{bm}
\usepackage{mathrsfs}
\usepackage{upgreek}

\title{\LARGE \bf Highly Dynamic Quadruped Locomotion via \\ Whole-Body Impulse Control and Model Predictive Control}
\author{Donghyun Kim$^{1}$, Jared Di Carlo$^{2}$, Benjamin Katz$^{1}$, Gerardo Bledt$^{1}$, and Sangbae Kim$^{1}$
\thanks{Authors are with the $^{1}$ Department of Mechanical Engineering at the Massachusetts Institute of Technology, and the $^{2}$ Department of Electrical Engineering and Computer Science at the Massachusetts Institute of Technology, Cambridge, MA, 02139, USA. Corresponding Author: {\tt \small robot.dhkim@gmail.com}}%
}

\begin{document}

\maketitle
\thispagestyle{empty}
\pagestyle{empty}

\begin{abstract}
Dynamic legged locomotion is a challenging topic because of the lack of established control schemes which can handle aerial phases, short stance times, and high-speed leg swings. In this paper, we propose a controller combining whole-body control (WBC) and model predictive control (MPC). In our framework, MPC finds an optimal reaction force profile over a longer time horizon with a simple model, and WBC computes joint torque, position, and velocity commands based on the reaction forces computed from MPC. Unlike existing WBCs, which attempt to track commanded body trajectories, our controller is focused more on the reaction force command, which allows it to accomplish high speed dynamic locomotion with aerial phases. The newly devised WBC is integrated with MPC and tested on the Mini-Cheetah quadruped robot. To demonstrate the robustness and versatility, the controller is tested on six different gaits in a number of different environments, including outdoors and on a treadmill, reaching a top speed of $3.7 \ \si{\meter\per\second}$.
\end{abstract}

\section{Introduction}
To fully exploit the hardware capability of legged systems, we need a controller that can address the challenging issues related to dynamic locomotion, such as body control during short stance periods, aerial phases, and high speed swing leg motion control. Several successful cases for both running bipeds \cite{Sreenath:2013en, Hubicki:2016vd} and quadrupeds~\cite{wildcat} have been presented, but they are either difficult to scale up to high degree-of-freedom systems~\cite{Sreenath:2013en} or heavily rely on specific system dynamics~\cite{Hubicki:2016vd} or are undocumented~\cite{wildcat}. Whole-body control (WBC) is a strong candidate as a dynamic motion controller because of its dynamically consistent formulation and general framework, which makes it easy to extend to various systems and tasks. However, existing WBCs focus on how to follow the given trajectory by manipulating contact forces, which makes it nontrivial to address motion involving frequent non-contact phases such as high speed running. 

\begin{figure}
    \centering
    \includegraphics[width=\columnwidth]{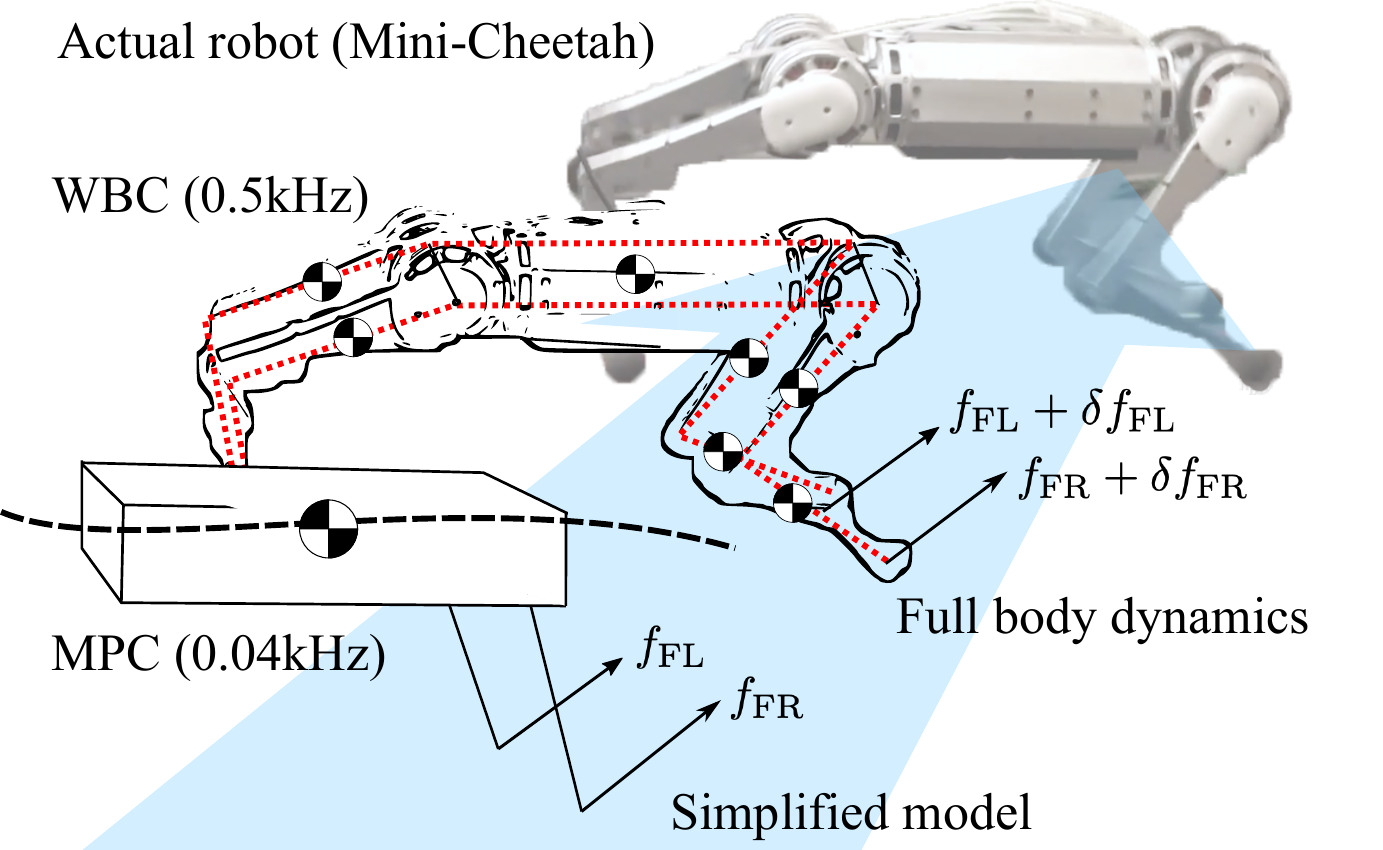}
    \caption{{\bf Control Architecture.} The proposed control architecture consists of two parts: Model predictive control and whole-body control. The reaction forces computed by MPC are modified by WBC to incorporate body stabilization and swing leg control. The final commands found in WBC are sent to the robot to perform dynamic locomotion.}
    \label{fig:my_label}
\end{figure}

\begin{figure*}
\centering
\includegraphics[width=2\columnwidth]{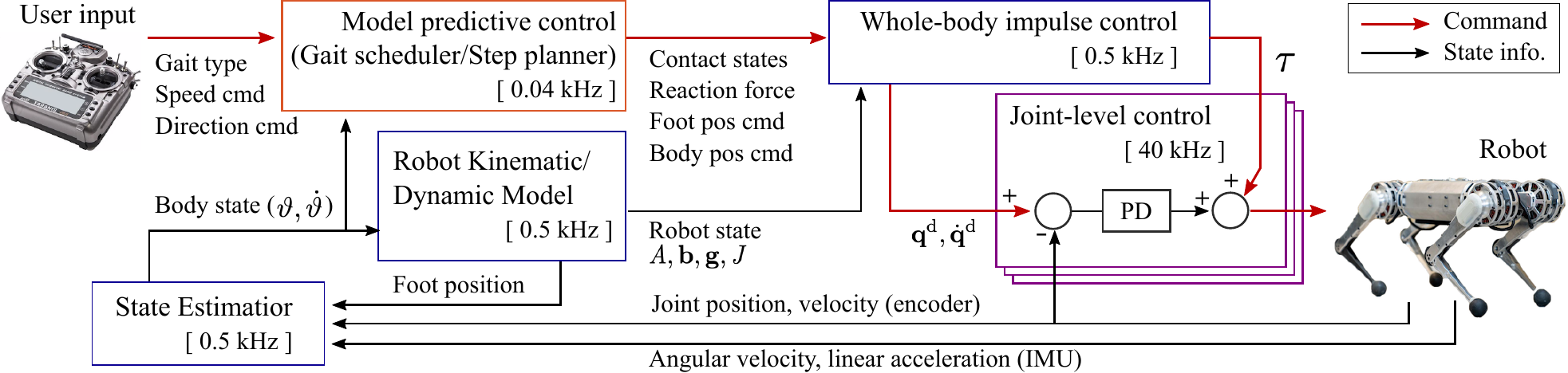}
\caption{{\bf Overall Control Framework.}  Using the user commanded gait type, speed, and direction from the RC-controller, the MPC computes desired reaction forces and foot/body position commands.  From these,  WBC computes joint torque, position, and velocity commands that are delivered to the each joint-level controller. Each component's update frequency is represented by the color of its box.}
\label{fig:ctrl_framework}
\end{figure*}

To tackle the issue, we formulate WBC to follow both the reaction force and body trajectory commands. The idea of reaction force tracking originates from the impulse planning used in Cheetah 2 \cite{Park:2017bi}, which demonstrates successful dynamic bounding and jumping. The underlying idea of \cite{Park:2017bi} is to plan reaction forces, which are impulses, rather than CoM trajectory, which is not practical to follow when the locomotion is extremely dynamic and has significant periods of under-actuation. In this paper, we embrace the impulse planning idea in WBC and formulate whole-body impulse control (WBIC) that can incorporate both body posture stabilization and reaction force execution. In terms of formulation, WBIC is not significantly different from the existing whole-body controllers \cite{Kim:2018iq,Henze:2016ct,righetti2012quadratic}, but the additional feature, which is an incorporation of pre-computed reaction forces by relaxing the floating base control, plays an important role in dynamic locomotion control. In our formulation, the WBIC is mostly used to track the ground reaction force profile rather than a body trajectory. 

To find the reaction force command, we utilize model predictive control (MPC). In our previous work, we demonstrated that convex MPC can perform various dynamic gaits at high speed on both Cheetah 3 \cite{DiCarlo:2018dz} and Mini-Cheetah \cite{Katz:2019vk}.
Utilization of MPC enhances the versatility of locomotion, enabling us to switch between various gaits by simply changing the contact sequence. However, using MPC with a simple model has a fundamental limitation in position control because of its low update frequency (40 Hz in our implementations) and model simplifications. WBC provides a solution to the MPC's limitation by running a high-frequency feedback loop while still accounting for full-body dynamics with contact. 

On the other hand, the prediction horizon of MPC compliments the WBC perfectly to fill in the WBC's limitation that it cannot consider more than a single time step ahead. This limited time horizon issue has been addressed in \cite{Neunert:iw, Koenemann:2015dx} which developed an MPC formulation using full-body dynamics. However, even their highly optimized solvers barely fit into a 200 Hz update frequency and the demonstrated results are not very dynamic compared to other controllers presented on the same robots. We believe that effective integration of MPC and WBC has a fundamental benefit over long-time horizon WBC. Moreover, by utilizing convex optimization, our implementation is not only fast, but also reliable because the solver does not get stuck in a strange local minimum.

To integrate the MPC with the WBC, we modified our previous WBC formulation \cite{Kim:2018iq} to be compatible with the MPC and to handle dynamic locomotion with aerial phases. However, the novelty of our work does not come from the integration of two methods, MPC and WBC, since the incorporation of WBC with a high-level trajectory generator such as a motion optimizer \cite{Feng:2015ixa}, locomotion planner \cite{Koolen:2016ci, Kim:2019vy}, or MPC \cite{Grandia:2019fc} is an established control framework. Our contribution lies in the method of how we utilize the results of MPC in the WBC. In our formulation, we use reaction forces computed by MPC as desired reaction forces in our WBC rather than attempting to track body trajectories computed by MPC. This is different from \cite{Grandia:2019fc} that forces the robot to follow the CoM trajectory found by MPC and uses the reaction forces found by MPC only for regulating internal forces. However, attempting to track a CoM trajectory with a WBC will not be effective for gaits like galloping, where the CoM is never controllable. Our WBIC controls the body posture and swing foot but with a relaxation variable that allows the floating base movement to be different from the commanded trajectory. By doing so, WBC can perform behaviors with uncontrollable center of mass (CoM), movement such as jumping or bounding, by controlling the reaction forces found by MPC.

Integrating WBC and MPC makes our controller versatile and robust. Versatile means that, in our controller, changing behavior or adding more tasks (e.g. manipulation if there are additional limbs) can be done by simply re-configuring the desired motions. For example, selecting different gaits can be done by changing the footstep pattern and timing in our formulation, and the low-level details such as swing foot control and body stabilization are automatically accomplished by our control scheme. Our controller will automatically handle gaits with significant flight periods, such as hopping, pronking, or bounding, without the need for manually planning around periods of flight or underactuation.

``Robust'' means that our controller is reliable enough to maintain the robot's balance while executing a locomotion command, even in the presence of large disturbances. More importantly, the proposed method can be implemented and demonstrates highly dynamic locomotion in a real robot. In our experiments with the Mini-Cheetah robot, we accomplished a $3.7\ \si{\meter\per\second}$ running speed, corresponding to a Froude number of 4.65. Most documented legged robots' Froude numbers are smaller than 2 \cite{Hubicki:2016vd, Hwangbo:2019dj}, except Cheetah 2 demonstrated $6.4\si{\meter\per\second}$ bounding (Froude number 7.1) \cite{Park:2017bi}. To the best of our knowledge, our results demonstrate some of the most agile locomotion of any quadruped robot.

 Additionally, in order to successfully implement the algorithm on hardware, we developed an efficient dynamics engine including rotor dynamics \cite{Jain:2010ek} and linear algebra optimizations through template formatting. These techniques enable the MPC to run at 30 Hz and WBIC to run at 500 Hz on the low power computer installed in the Mini-Cheetah robot. In summary, the major contributions of this paper are two folds: 1) developing a versatile and robust control scheme for highly dynamic locomotion of quadruped robots, and 2) demonstrating high speed running and various gaits in real hardware, on the Mini-Cheetah robot.

\section{Hybrid Control Architecture}

The key idea of our method is to reduce complexity by separating locomotion control into two simpler controllers. This first controller finds optimal reaction force profiles along a one full gait cycle of locomotion using an MPC with the following simple lumped mass model:

\begin{align}
\label{eq:simple_lin}
    &m \ddot{\mathbf{p}} = \sum_{i=1}^{n_c} \mathbf{f}_i - \mathbf{c}_g, \\
    \label{eq:simple_ori}
    &\frac{d}{dt}\left(\bm{I}\bm{\omega}\right)=\sum_{i=1}^{n_c}\mathbf{r}_i\times \mathbf{f}_i,
\end{align}
where $\mathbf{p}$, $\mathbf{f}_i$, and $\mathbf{c}_g$ are three dimensional vectors representing the robot's position, reaction force, and gravitational acceleration with respect to the global frame. $m$ is the robot's body mass and $n_c$ is the number of contacts. $\bm{I} \in \mathbb{R}^{3\times3}$ is the rotational inertia tensor and $\bm{\omega}$ is the angular velocity of the body. $\mathbf{r}_i$ is the position of $i$-th contact point with respect to the CoM of the robot, which is equivalent to the moment arm of the contact force. 

In the second process, we use WBIC to achieve high bandwidth control through the use of full-body dynamics and high frequency feedback control. This better dynamics model will determine more accurate torque commands than the lumped mass model.  The multi-body dynamics can be written as

\begin{equation} \label{eq:full_dyn}
    \bm{A}
    \begin{pmatrix}
    \ddot{\mathbf{q}}_f \\
    \ddot{\mathbf{q}}_j
    \end{pmatrix}+ \mathbf{b}+\mathbf{g} = 
    \begin{pmatrix}
    \mathbf{0}_6 \\ \bm{\tau}
    \end{pmatrix}
    +\bm{J}_c^{\top}\mathbf{f}_r,
\end{equation}
where $\bm{A}$, $\mathbf{b}$, $\mathbf{g}$, $\bm{\tau}$, $\mathbf{f}_r$, and $\bm{J}_c$ are the generalized mass matrix, Coriolis force, gravitation force, joint torque, augmented reaction force and contact Jacobian, respectively. $\ddot{\mathbf{q}}_f\in \mathbb{R}^6$ is the acceleration of the floating base and $\ddot{\mathbf{q}}_j\in\mathbb{R}^{n_j}$ is the vector of joint accelerations, where $n_j$ is the number of joints. We use $\mathbf{0}_6$ to represent a 6 dimensional zero vector and $\mathbf{0}_{n}$ to represent a $n$-dimensional zero vector in this paper. 

Our hybrid control scheme combining these two controllers is described in Fig.~\ref{fig:ctrl_framework}. Beside MPC and WBIC, we use a Kalman Filter-based state estimator to compute global body position and velocity based on kinematics and acceleration data. We also developed a custom dynamics engine that efficiently includes the effects of rotor dynamics on the mass matrix and Coriolis vector. Although the the state estimator and dynamics engine contributed significantly to the locomotion performance, we will not explain them in detail since they are out of the scope of this paper.

\section{Model Predictive Control}
The objective of our MPC to find reaction forces which make the lumped mass follow the given trajectory. In the optimization process of the MPC, a contact sequence is predefined by the gait scheduler and step planner. This keeps the formulation convex, meaining the optimization problem is both fast to solve and can be always solved to a unique global minimum.  This is not always obtainable in nonlinear optimization.

\subsection{MPC Formulation}
Even the simple lumped mass model is not completely linear due to the cross product term for the moment arm and the orientation dynamics. To accomplish a convex MPC formulation, we applied three simplifications~\cite{DiCarlo:2018dz}. The first assumption is that the roll and pitch angles are small. Based on this assumption, we can simplify the coordinate transformation as follows. 
\begin{align}
     \dot{\bm{\Theta}}&\approx \bm{R}_z(\psi)\bm{\omega}, \\
    _\mathcal{G}\bm{I} &\approx \bm{R}_z(\psi) _\mathcal{B}\bm{I} \bm{R}_z(\psi)^{\top},
\end{align}
where $\dot{\bm{\Theta}} = \begin{bmatrix} \dot{\phi} & \dot{\theta} & \dot{\psi} \end{bmatrix}^{\top}$ is angular velocity of the body with roll ($\phi$), pitch ($\theta$), and yaw ($\psi$) Euler angle representation. $\bm{R}_z(\psi)$ is a rotation matrix translating angular velocity in the global frame, $\bm{\omega}$, to the local (body) coordinate. $_\mathcal{G}\bm{I}$ and $_\mathcal{B}\bm{I}$ are the inertia tensor seen from the global and local (body) frame, respectively.

The second assumption we made is that states are close to the commanded trajectory. Based on this assumption, we create a time-varying linearization of the dynamics using commanded $\psi$ in a rotational matrix, $\bm{R}_z(\psi)$, and set the moment arm in Eq.~\eqref{eq:simple_ori} with the predetermined one from the commanded trajectory and step locations. The last assumption is that the pitch and roll velocities are small and off-diagonal terms of the inertia tensor are also small. With the assumption, we approximate Eq.~\eqref{eq:simple_ori} with the following:
\begin{equation}
    \frac{d}{dt}\left(\bm{I}\bm{\omega} \right) = 
    \bm{I}\dot{\bm{\omega}} + \bm{\omega}\times\left( \bm{I} \bm{\omega} \right) \approx \bm{I}\dot{\bm{\omega}}.
\end{equation}

With the above three simplifications, the discrete dynamics of the system can be expressed as

\begin{equation}
    \mathbf{x}(k+1) = \bm{A}_k\mathbf{x}(k) + \bm{B}_k\hat{\mathbf{f}} (k)+ \hat{\mathbf{g}},
\end{equation}
where,
\begin{equation}
\begin{split}
\mathbf{x} &= \begin{bmatrix}
\bm{\Theta}^{\top} & \mathbf{p}^{\top} & \bm{\omega}^{\top} & \dot{\mathbf{p}}^{\top} 
\end{bmatrix}^{\top}, \\[1.5mm]
    \hat{\mathbf{f}} &= \begin{bmatrix}
    \mathbf{f}_1 & \cdots & \mathbf{f}_n
    \end{bmatrix}^{\top},\\[1.5mm]
    \hat{\mathbf{g}} &= \begin{bmatrix}
    \bm{0}_{1\times3} & \bm{0}_{1\times3} & \bm{0}_{1\times3}&\mathbf{g}^{\top}
    \end{bmatrix}^{\top},
    \end{split}
\end{equation}
\begin{equation}
\begin{split}
    \bm{A} &= \begin{bmatrix}
    \bm{1}_{3\times3} & \bm{0}_{3\times3} & \bm{R}_z(\psi_k) \Delta t & \bm{0}_{3\times3} \\[1mm]
    \bm{0}_{3\times3} & \bm{1}_{3\times3} & \bm{0}_{3\times3} &
    \bm{1}_{3\times3} \Delta t \\[1mm]
    \bm{0}_{3\times3} & \bm{0}_{3\times3} & \bm{1}_{3\times3} &
    \bm{0}_{3\times3} \\[1mm]
    \bm{0}_{3\times3} & \bm{0}_{3\times3} & \bm{0}_{3\times3} &
    \bm{1}_{3\times3}
    \end{bmatrix}, \\[1.5mm]
    \bm{B} &= \begin{bmatrix}
    \bm{0}_{3\times3} & \cdots &  \bm{0}_{3\times3} \\[1mm]
    \bm{0}_{3\times3} & \cdots &  \bm{0}_{3\times3} \\[1mm]
_\mathcal{G}\bm{I}^{-1}[\mathbf{r}_1]_{\times} \Delta t& \cdots & _\mathcal{G}\bm{I}^{-1}[\mathbf{r}_n]_{\times} \Delta t\\[1mm]
    \bm{1}_{3\times3} \Delta t/m& \cdots & \bm{1}_{3\times3}
\Delta t/m    \end{bmatrix}.
\end{split}
\end{equation}
We use the formulation described in \cite{DiCarlo:2018dz} to construct a QP which minimizes
\begin{equation}\label{eq:mpc_qp}
    \min_{\mathbf{x}, \mathbf{f}} \sum_{k=0}^{m}||\mathbf{x}(k+1) - \mathbf{x}^{\rm ref}(k+1)||_{\bm{Q}} + ||\mathbf{f}(k)||_{\bm{R}}
\end{equation}
subject to dynamics and initial condition constraints.  Additionally, friction cones are approximated by the following ground reaction force constraints
\begin{equation}
\label{eq:contact_constraint}
\mid f_x \mid \leq \mu f_z, \quad \mid f_y \mid \leq  \mu f_z,
\quad f_z > 0.
\end{equation}

To reduce the size of the problem, we eliminate variables which correspond to ground reaction forces for feet which are not touching the ground.  This reduces the size of both the cost and constraint matrix, and in practice gives us a speed up of over 10 times. Since all contacts in our problem are point contacts, we do not need to account for the non-flip condition that is described in \cite{Caron:2015tf}. 

\subsection{Gait Scheduler}
We use a periodic phase-based gait scheduler introduced in \cite{8460904} that needs only two parameters to specify a gait type; phase offset and stance period for each foot. Since most gaits are basically periodic, proper spacing of swing and stance period in one cycle is enough to define different gaits. We can easily change the gait frequency by changing the cycle duration. Because gaits are defined by a portion of stance/swing period in the cycle, the gait types are maintained even when the gait frequency changes. 

\subsection{Foot Step Planner}
We choose the upcoming foot step location with the following equation:

\begin{equation}
    \mathbf{r}^{\rm cmd}_i = \mathbf{p}_{{\rm shoulder},i }+ \mathbf{p}_{\rm symmetry} + \mathbf{p}_{\rm centrifugal},
\end{equation}
where, 
\begin{align}
\label{eq:shoulder}
    &\mathbf{p}_{{\rm shoulder}, i} = \mathbf{p}_k + \bm{R}_z\left( \psi_k\right)\mathbf{l}_i, \\[2mm]
    &\mathbf{p}_{\rm symmetry} =  \frac{t_{\rm stance}}{2}\mathbf{v} + k \left(\mathbf{v} - \mathbf{v}^{\rm cmd} \right),\\
      & \mathbf{p}_{\rm centrifugal} = \frac{1}{2}\sqrt{\frac{h}{g}} \mathbf{v}\times\bm{\omega}^{\rm cmd},
\end{align}
In Eq.~\eqref{eq:shoulder}, $\mathbf{p}_k$ is the body position at the $k$-th timestep and $\mathbf{l}_i$ is $i$-th leg shoulder location with respect to the body's local frame. Therefore, $\mathbf{p}_{{\rm shoulder}, i}$ is the $i$-th shoulder location with respect to the global frame. $\mathbf{p}_{\rm symmetry}$ is a so-called Raibert heuristic \cite{Raibert:1984ej} that forces the leg's landing angle and leaving angle be identical if the robot is traveling at the commanded velocity. In our setup, we use 0.03 for the feedback gain, $k$. 

\section{Whole-Body Impulse Control}
Using the reaction forces found by the MPC, the WBIC computes joint position, velocity, and torque commands. For joint position, velocity, and acceleration computation, we utilize an inverse kinematics algorithm that strictly holds task priority. To compute the torque command, we use quadratic programming to find the reaction forces, reducing the both errors in acceleration command tracking and reaction force command tracking while satisfying inequality constraints on the resultant reaction forces. 

Joint position and velocity commands are used to stabilize a posture through joint-level position controllers. In addition to the torque command, which is the final output of common WBC, WBIC computes a desired joint position and velocity. Utilizing joint position feedback is beneficial for dynamic locomotion control because of its collocated control input and high frequency update. In the case of Mini-Cheetah, the frequency of joint PD control is $40\si{\kilo\hertz}$, which is 80 times faster than the high-level full body control. The effectiveness of joint position feedback in whole-body control is well demonstrated in the author's previous work, passive-ankle biped walking~\cite{Kim:2019vy}. In \cite{Kim:2019vy}, the robot accomplish the stable body posture control and accurate swing foot control by utilizing joint position control. WBIC uses similar strategy to obtain reliable motion stabilization.

Although we take a similar approach to the author's previous work \cite{Kim:2019vy}, we use operational space control instead of configuration impedance control. We made this change based on difference between actuators used in each robot: proprioceptive actuators in Mini-Cheetah and series elastic actuators in Mercury. Since the proprioceptive actuator is highly backdrivable and provides open-loop joint torque control through the motor current control, it is more effective to directly control operational impedance rather than relying on joint impedance control that can delay the operational space control. The following sections explain the formulation of WBIC in detail. 

\subsection{Prioritized Task Execution}
\label{sec:null_space}
To execute prioritized tasks, we utilize a null-space projection technique, which enables a strict task hierarchy in a computationally efficient way. When $\mathbf{q} = \begin{bmatrix}\mathbf{q}_f^{\top} & \mathbf{q}_j^{\top}\end{bmatrix}^{\top}$ is a vector representing full configuration space, the iteration rules are the following. 
\begin{align}
\label{eq:delta_q}
&\Delta \mathbf{q}_i = \Delta \mathbf{q}_{i-1} + {\bm{J}_{i| pre}}^{\dagger}\left( \mathbf{e}_i - \bm{J}_i\Delta \mathbf{q}_{i-1} \right), \\[2mm]
\label{eq:qdot_cmd}
&\dot{\mathbf{q}}_i^{\rm cmd} = \dot{\mathbf{q}}_{i-1}^{\rm cmd} + {\bm{J}_{i|pre}}^{\dagger}\left( \dot{\mathbf{x}}^{\rm des}_i - \bm{J}_i\dot{\mathbf{q}}_{i-1}^{\rm cmd}\right), \\[2mm]
\label{eq:qddot_cmd}
&\ddot{\mathbf{q}}_{i}^{\rm cmd} = \ddot{\mathbf{q}}_{i-1}^{\rm cmd} + \overline{\bm{J}_{i|pre}^{\rm dyn}}\left( \ddot{\mathbf{x}}_{i}^{\rm cmd} - \dot{\bm{J}}_i\dot{\mathbf{q}} - \bm{J}_{i} \ddot{\mathbf{q}}_{i-1}^{\rm cmd} \right),
\end{align}
where
\begin{align}
\begin{split}
	&\bm{J}_{i|pre} = \bm{J}_{i} \bm{N}_{i-1}, \\
    &\bm{N}_{i-1} = \bm{N}_{0}\bm{N}_{1|0} \cdots \bm{N}_{i-1|i-2},
\end{split}\\[2mm]
\begin{split}\label{eq:null_space_matrix}
    &\bm{N}_{0} = \bm{I} - \bm{J}^{\dagger}_{c}\bm{J}_c,\\
    &\bm{N}_{i|i-1} = \bm{I} - {\bm{J}_{i|i-1}}^{\dagger} \bm{J}_{i|i-1}.
\end{split}
\end{align}
Here, $i\geq 1$, and 
\begin{equation}
    \begin{split}
        \Delta \mathbf{q}_0,\  \dot{\mathbf{q}}_0^{\rm cmd} &= \mathbf{0} ,\\
        \ddot{\mathbf{q}}_0^{\rm cmd} &= \overline{\bm{J}_c^{\rm dyn}}(-\bm{J}_c\dot{\mathbf{q}}).
    \end{split}
\end{equation}
$\mathbf{e}_i$ is the position error defined by $\mathbf{x}_i^{\rm des} - \mathbf{x}_i$ and $\ddot{\mathbf{x}}^{\rm cmd}_i$ is the acceleration command of $i$-th task defined by
\begin{equation}
    \ddot{\mathbf{x}}^{\rm cmd}_i = \ddot{\mathbf{x}}^{\rm des} + \bm{K}_p \left(\mathbf{x}^{\rm des}_i - \mathbf{x}_i\right) + \bm{K}_d\left(\dot{\mathbf{x}}^{\rm des} - \dot{\mathbf{x}}\right),
\end{equation}
where $\bm{K}_p$ and $\bm{K}_d$ are position and velocity feedback gains, respectively. Note that there is no feedback gain in Eq.~\eqref{eq:delta_q}, which can be interpreted as using unity gains. $\bm{J}_{i|pre}$ is the projection of the $i$-th task Jacobian into the null space of the prior tasks. $\bm{J}_c$ is a contact Jacobian, which is equivalent to the $\bm{J}_c$ in Eq.~\eqref{eq:full_dyn}. We use two types of pseudo-inverses; one is an SVD-based pseudo-inverse denoted by $\{\cdot \}^{\dagger}$ and the other is dynamically consistent pseudo-inverse defined by
\begin{equation}
    \overline{\bm{J}} = \bm{A}^{-1}\bm{J}^{\top}\left( \bm{J} \bm{A}^{-1} \bm{J}^{\top} \right)^{-1}.
\end{equation}
When computing acceleration (Eq.~\eqref{eq:qddot_cmd}), we use the dynamically consistent pseudo-inverse. Therefore, the projected Jacobian, $\bm{J}_{i|pre}^{\rm dyn}$, is different from the ones used in kinematics computations. The dynamically consistent Jacobians use $\overline{\{\cdot \}}$ instead of $\{\cdot \}^{\dagger}$ in the null space matrix computation (Eq.~\eqref{eq:null_space_matrix}).

Eq.~\eqref{eq:delta_q} and \eqref{eq:qdot_cmd} are used to find desired joint position and velocity for joint PD controller, respectively. We compute the desired joint position by adding the joint position portion of Eq.~\eqref{eq:delta_q} to the measured joint position,
\begin{equation}
    \mathbf{q}_j^{\rm cmd} = \mathbf{q}_j + \Delta \mathbf{q}_j.
\end{equation}
The computed joint commands, $\mathbf{q}_j^{\rm cmd}$ and $\dot{\mathbf{q}}_j^{\rm cmd}$, are sent to the joint-level PD controller, and the acceleration commands $\ddot{\mathbf{q}}^{\rm cmd}$ are delivered to the QP optimization to find a torque command.  

\subsection{Quadratic Programming}
We compute the final reaction force with the acceleration command found in the previous step and the reaction force obtained in MPC. For the optimization, we use the open-source QP solver \cite{Goldfarb:1983ik} that is efficient for small problems. The formulation of our QP problem is

\begin{equation} \label{eq:qp_cost}
\min_{\bm{\delta}_{\mathbf{f}_r}, \bm{\delta}_{f}}\quad \bm{\delta}_{\mathbf{f}_r}^{\top} \bm{Q}_1 \bm{\delta}_{\mathbf{f}_r} + \bm{\delta}_{f}^{\top}\bm{Q}_2\bm{\delta}_{f}\vspace{0.7mm} \\
 \vspace{-4mm}
\end{equation}
\begin{align*}
\text{s.t.} & \\
\tag{floating base dyn.}
&\bm{S}_f 
\left(
\bm{A} \ddot{\mathbf{q}} + \mathbf{b} + \mathbf{g}
\right) 
= \bm{S}_f \bm{J}_{c}^{\top} \mathbf{f}_r \\
\tag{acceleration}
&\ddot{\mathbf{q}} = \ddot{\mathbf{q}}^{\rm cmd} +  \begin{bmatrix}
\bm{\delta}_{f} \\ \mathbf{0}_{n_j}
\end{bmatrix} \\
\tag{reaction forces}
&\mathbf{f}_r 
= \mathbf{f}_r^{\rm MPC} + \bm{\delta}_{\mathbf{f}_r}\\
\tag{contact force constraints}
 & \bm{W} \mathbf{f}_r \geq \mathbf{0},
\end{align*}
where $\mathbf{f}_r^{\rm MPC}$ and $\bm{S}_f$ are reaction forces computed by the MPC and the floating base selection matrix, respectively. $\bm{J}_c$ and $\bm{W}$ are the augmented contact Jacobian and contact constraint matrix that are equivalent to the terms used in Eq.~\eqref{eq:full_dyn} and Eq.~\eqref{eq:mpc_qp}, respectively. $\bm{\delta}_{f}$ and $\bm{\delta}_{\mathbf{f}_r}$ are relaxation variables for the floating base acceleration and reaction forces. 

Because of the relaxation of floating base acceleration, task accelerations can differ from the ones computed in Section.~\ref{sec:null_space}. The difference is intended to allow for the base to be uncontrolled during a flight phase, but has a risk to introduce tracking errors to other tasks. We ignore the effect on the other tasks because changing task commands as a response of the unpredictable floating base motion is not desirable in the real hardware control. For example, when a robot jumps, we need to address a floating base acceleration in the computation of foot acceleration command to keep the strict task priority. However, in most real hardware experiments, considering a floating base acceleration incorporating with complex dynamics and gravitational force does not help enhancing the swing foot control. Therefore, we simply relax the floating base dynamics rather than strictly govern the prioritized task execution by utilizing complex algorithms such as hierarchical quadratic programming \cite{escande2014hierarchical}. 

The last step of WBIC is to compute a torque command from the 
reaction forces, $\mathbf{f}_r$, and the configuration space acceleration, $\ddot{\mathbf{q}}$. By plugging these two terms into Eq.~\eqref{eq:full_dyn}, we obtain 
\begin{equation}
\begin{bmatrix}
\bm{\tau}_f \\
\bm{\tau}_{j}
\end{bmatrix} =
\bm{A}\ddot{\mathbf{q}}  + \mathbf{b} + \mathbf{g} - \bm{J}_c^{\top}\mathbf{f}_r.
\end{equation}
Since all terms on the right hand side are known, we can easily solve it and obtain the joint torque command, $\bm{\tau}_j$.

\section{Experimental Results}
\begin{figure}
\centering
\includegraphics[width=1.0\columnwidth]{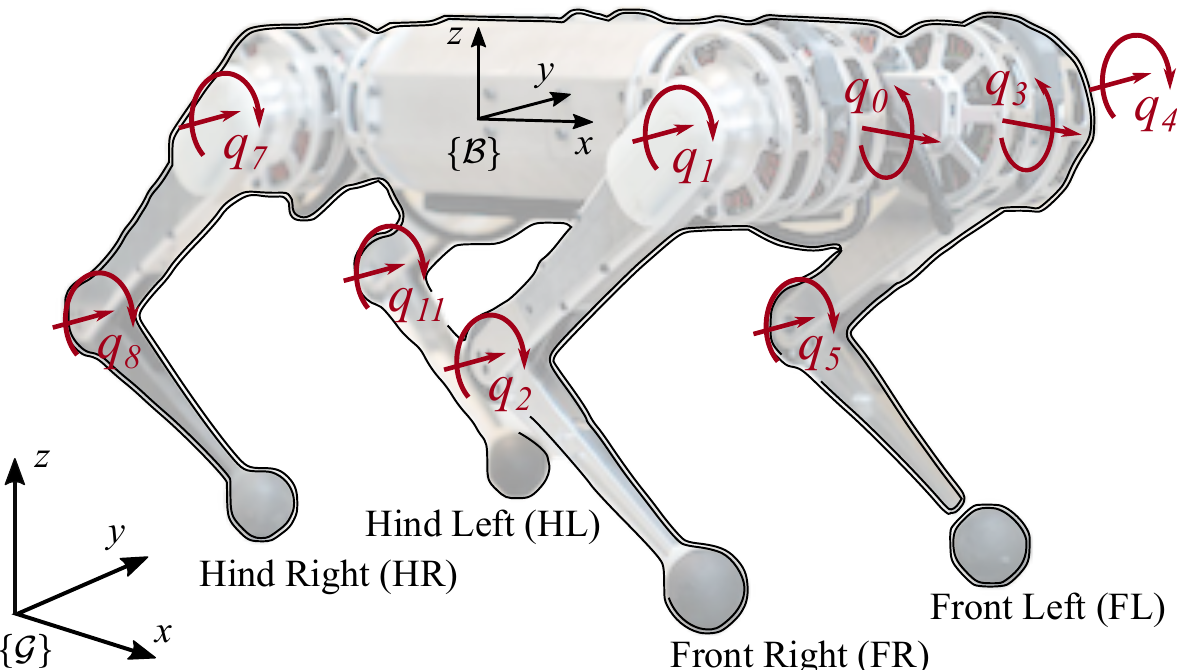}
\caption{{\bf Configuration of Mini-Cheetah.} Mini-cheetah uses 12 proprioceptive actuators to control 4 limbs. The numbering of joints starts from right front abduction/adduction joint and progresses to hip and knee flexion/extension joints.}
\label{fig:mini_cheetah_config}
\end{figure}

\begin{table}
\centering
\begin{tabular}{>{\centering}m{0.27\columnwidth} %
                >{\centering}m{0.27\columnwidth} %
                >{\centering}m{0.27\columnwidth} %
                @{}m{0pt}@{}}
\specialrule{1.5pt}{1pt}{1pt}
{Task}
&{$\mathbf{k}_p\  (\si{\second}^{-2})$}
&{$\mathbf{k}_d\  (\si{\second}^{-1})$}
&\\[2mm] 
\hline
\hline 
Body Orientation & 
$\begin{bmatrix}
100 & 100 & 100
\end{bmatrix}^{\top}$ & 
$\begin{bmatrix}
10 & 10 & 10
\end{bmatrix}^{\top}$
& \\[2mm]
\hline 
Body Position & 
$\begin{bmatrix}
100 & 100 & 100
\end{bmatrix}^{\top}$ & 
$\begin{bmatrix}
10 & 10 & 10
\end{bmatrix}^{\top}$
& \\[2mm]
\hline 
Foot Position & 
$\begin{bmatrix}
100 & 100 & 100
\end{bmatrix}^{\top}$ & 
$\begin{bmatrix}
10 & 10 & 10
\end{bmatrix}^{\top}$
& \\[2mm]
\hline
\end{tabular}
\caption{Task and Gain Setup}
\label{tb:task_setup}
\vspace{-3.5mm}
\end{table}

The Mini-Cheetah robot \cite{Katz:2018vn} is used to verify the effectiveness of the proposed controller. The task setup and feedback gains of WBIC used in the experiments are summarized in Table.~\ref{tb:task_setup}. The tasks are listed in order of priority. In the QP problem, the weight for the reaction force ($\bm{Q}_1$) is 1 and the weight for the floating base control ($\bm{Q}_2$) is 0.1. Every joint feedback controller shares the same gains, $k_p = 3\  \si{\newton\meter\per\radian}$ and $k_d = 0.3\  \si{\newton\meter\second\per\radian}$, except abduction joints, which have a higher derivative gain, 1. The same parameter and task setup is used over all tests. The labels of feet and joints used in the description coincide with the labels depicted in Fig.~\ref{fig:mini_cheetah_config}. A video recording of the experiments can be found in \url{https://youtu.be/6JlVol3eyNI}.

\subsection{High Speed Running}

\begin{figure*}
    \centering
    \includegraphics[width=2\columnwidth]{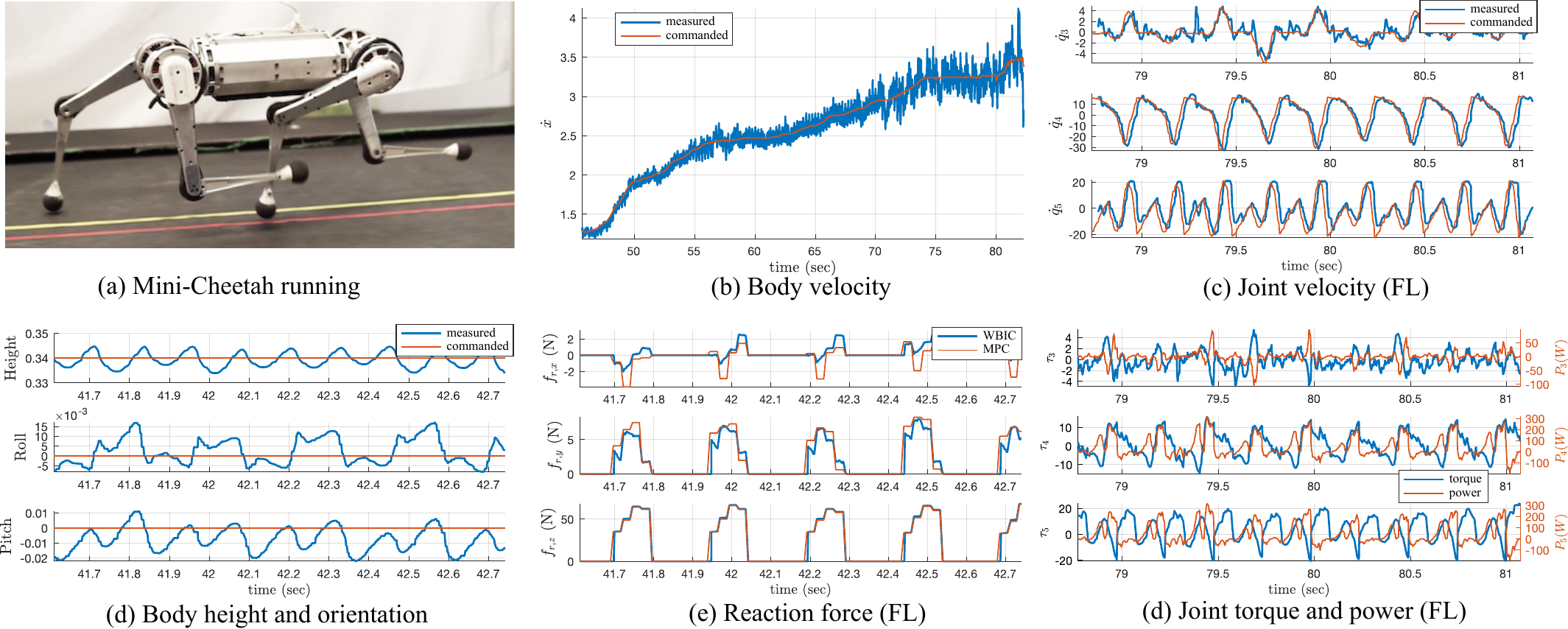}
    \caption{{\bf Running Experiment.} Mini-Cheetah trots at $3.7\si{\meter\per\second}$. (b) The observed maximum speed is $4 \ \si{\meter\per\second}$ but the robot quickly lost its balance after reaching this speed. The highest stable speed reached was is $3.7\  \si{\meter\per\second}$. (c) and (d) show the joint velocity, torque, and power. We only include data from three joints (abduction, hip, and knee) from the front left leg, but the other legs are similar. From the observed maximum velocity, torque, and power, we can see that our controller utilizes maximum hardware capacity of Mini-Cheetah.
    (d) and (e) show how the trajectory and reaction forces tracking work together to accomplish dynamic running. The robot's height goes up and down around the constant height command while making jumps and landing, which is accomplished by following the vertical reaction force command ($f_z$). (e) Reaction forces computed by MPC update only 4 times during a stance period, but WBIC computes the forces every $2\ \si{\milli\second}$ and makes a modification from the force commands to control body posture and swing feet.}
    \label{fig:exp_run}
\end{figure*}

In our previous work~\cite{Katz:2019vk}, Mini-Cheetah accomplished the running speed of $2.45\ \si{\meter\per\second}$ in maximum. Considering the size of the robot, the speed is comparable to the state-of-the-art such as WildCat ($8\ \si{\meter\per\second}$) and Cheetah 2 ($6.4\ \si{\meter\per\second}$), but was limited by the stability of the previous MPC-only controller, rather than the Mini-Cheetah hardware. Our new controller combining MPC and WBIC achieves a maximum forward velocity of $3.7\ \si{\meter\per\second}$, which is one of the fastest untethered quadruped robot running speed. 

For high speed running, we add one more feature in the step location algorithm. As the speed goes up, Mini-Cheetah narrows the step-width of the front feet and widens the hind feet to avoid collisions between the front and hind legs. Fig.~\ref{fig:exp_run}(a) shows that the right front leg and right hind legs are crossed but do not collide with each other because of the step-width adjustment. 

Fig.~\ref{fig:exp_run} summarizes the test results of a high speed running on a treadmill. The robot's velocity estimated by the local foot speed and accelerometer is presented in Fig.~\ref{fig:exp_run}(b). The maximum velocity we observed is $4\ \si{\meter\per\second}$, but the robot falls over right after the speed. Therefore it is more reasonable to see $3.7\ \si{\meter\per\second}$ as our record. Fig.~\ref{fig:exp_run}(c) and (d) show how each joint operates during the high speed running. The maximum joint velocity and torque at the hip joints are $34\ \si{\radian\per\second}$ and $25.5\ \si{\newton\meter}$, respectively. Considering that the maximum joint velocity and torque are $40\ \si{\radian\per\second}$ and $17\ \si{\newton\meter}$, we can conclude that the hardware capability is fully utilized in the test. The maximum capacity utilization is also confirmed by the power output presented in Fig.~\ref{fig:exp_run}(d), which records $280\ \si{\watt}$, which is larger than the maximum actuator power, $250\ \si{\watt}$, specified in \cite{Katz:2019vk}. Note that the torque presented in the data is not measured joint torque but commanded torque to motor controller. Therefore, it can have larger number than the actuator limit, and in that case when the commanded torque is larger than the limit, the actuator output is truncated by the maximum output torque. 

Fig.~\ref{fig:exp_run}(d) and (e) show how the trajectory and reaction force tracking cooperate together to perform high speed running including aerial phase. The command for body posture is constant, which is the same command as for MPC, but WBIC does not force the robot to exactly follow the commanded trajectory. Instead, WBIC finds the solution satisfying the reaction force commands computed by MPC while controlling the body posture and swing feet as much as possible. Therefore, the measured height and pitch in Fig.(d) goes up and down around the constant commands since the robot jumps and lands as MPC plans.

\subsection{Outdoor Test}
\begin{figure*}
    \centering
    \includegraphics[width=2\columnwidth]{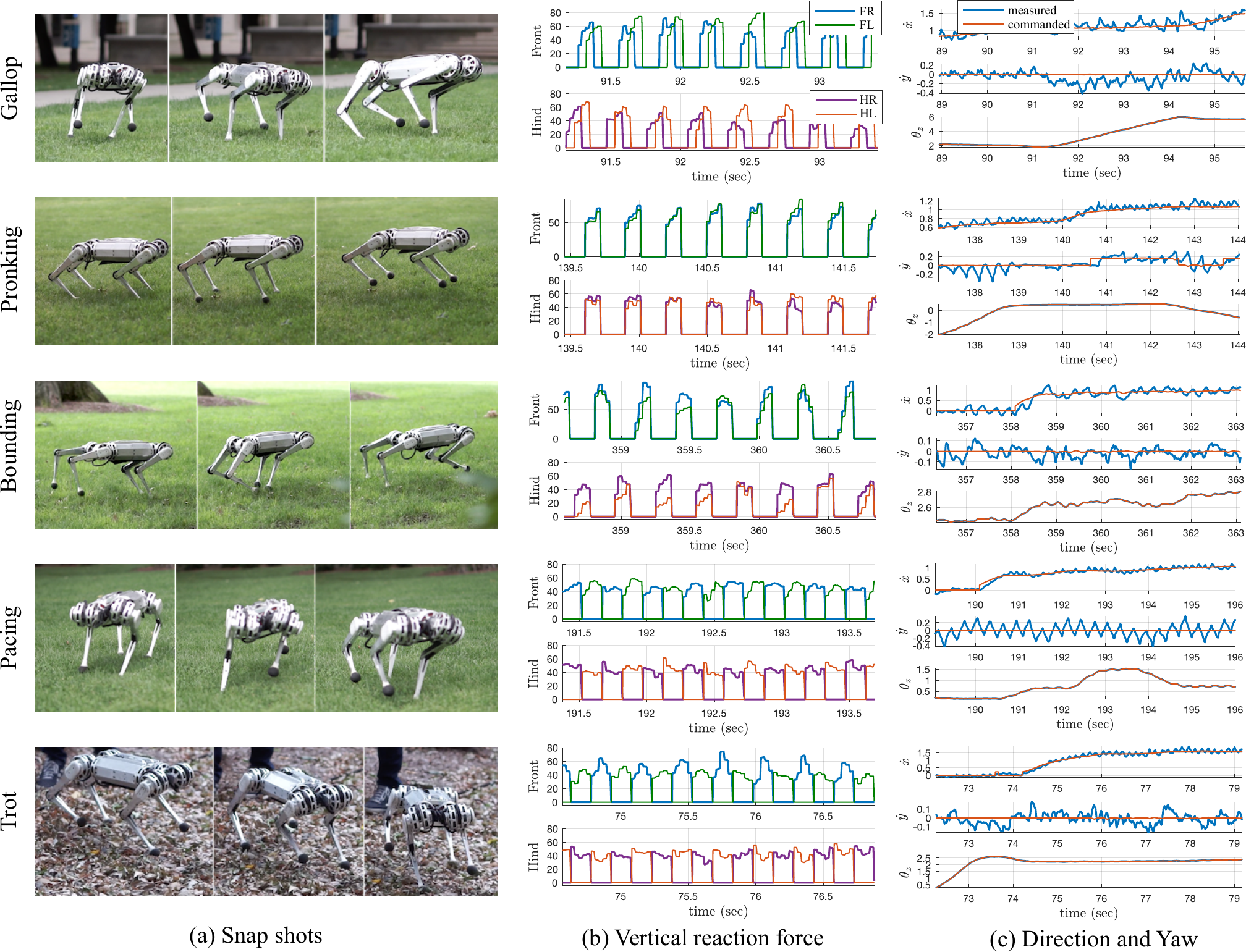}
    \caption{{\bf Demonstration of Various Gaits.} Our controller can demonstrate various dynamic running by switching the gait types. (b) The commanded vertical reaction forces are presented. (c) The commanded velocity and direction are presented with the measured velocity and yaw angle. The results show that the actual system follows the given commands well.} 
    \label{fig:gait}
\end{figure*}

Fig.~\ref{fig:gait} shows outdoor dynamic locomotion of various gait types. In the test, we use the same gains and weights over all gait types as we mentioned above. Over the entire outdoor test, the only inputs that we provide to the controller are gait type, running speed and direction, and all of them are transferred to the robot through a remote controller. Fig.~\ref{fig:gait} is a summary of the outdoor test. As the results show, our controller is capable of performing the various gaits in a stable and robust way. The robot follows the commanded velocity and direction well as can be seen in Fig.~\ref{fig:gait}(c). Mini-Cheetah can run at speeds over $1\ \si{\meter\per\second}$ with every gait type. Fig.~\ref{fig:gait}(b) shows that our controller commands  correct reaction force profiles as a desired gait type is changed.

The controller is also robust to different terrain types. The green field was wet and slippery because of the rain from the day before the test, but Mini-Cheetah stably keeps its body posture and accurately controls the swing legs. The robot can also run over the gravel terrain, which has significant roughness and rolling elements that cause body and step control to be difficult. 

\section{Conclusion and Discussion}
In this paper, we propose a new control scheme that can accomplish various dynamic locomotion of a quadruped robot with minimal user intervention. The effective combination of MPC and WBC is verified by the experimental results on the Mini-Cheetah robot. We accomplish high speed running up to $3.7\ \si{\meter\per\second}$ by fully utilizing the hardware capability. In the outdoor tests, Mini-Cheetah demonstrates various gaits with speed over $1\ \si{\meter\per\second}$ and push-recovery on rough terrain.

The immediate next step will be an implementation of the identical control framework to Cheetah 3 robot. We expect that no additional work is required for the implementation since the formulation is independent from the system parameters. Simple switch of system dynamics will enable high performance locomotion on Cheetah 3. A more interesting extension will be adding manipulation tasks on the cheetah robots with an additional manipulator. Although we do not mention various tasks except locomotion, WBC is naturally applicable to loco-manipulation tasks. With a small addition to the current scheme, our controller will be able to manage both locomotion and manipulation.

The proposed control scheme does not have a limitation not only in the task types but also for a topology of the target system. Therefore, we can apply this method to biped locomotion. We plan to extend our controller to biped walking and running in the near future with small additions such as a new contact sequence and step location planner.

\section*{Acknowledgments}
This work was supported by the National Science Foundation [NSF- IIS-1350879], Naver Labs, and the Air Force Office of Scientific Research.

\bibliographystyle{IEEEtran}
\bibliography{mpc_wbc}

\end{document}